\documentclass{article}
\usepackage{ijcai21}

\usepackage{times}
\usepackage{soul}
\usepackage{url}
\usepackage[hidelinks]{hyperref}
\usepackage[utf8]{inputenc}
\usepackage[small]{caption}
\usepackage{graphicx}
\usepackage{amsmath}
\usepackage{amsthm}
\usepackage{booktabs}
\usepackage{algorithm}
\usepackage{algorithmic}

\usepackage{amsfonts} 
\DeclareMathOperator*{\argmax}{arg\,max}
\DeclareMathOperator*{\argmin}{arg\,min}
\usepackage{subfigure}
\usepackage{booktabs}
\usepackage{enumitem}
\usepackage{bbm}
\usepackage{lineno}
\DeclareMathOperator{\relu}{ReLU}
\DeclareMathOperator{\diag}{diag}
\DeclareMathOperator{\erf}{erf}

\title{GUAP: Graph Universal Attack Through Adversarial Patching}

\author{
Xiao Zang$^1$
\and
Jie Chen$^2$\footnote{Contact Author}\And
Bo Yuan$^1$
\affiliations
$^1$Department of Electrical and Computer Engineering, Rutgers University\\
$^2$MIT-IBM Watson AI Lab, IBM Research
\emails
xz514@scarletmail.rutgers.edu,
chenjie@us.ibm.com,
 bo.yuan@soe.rutgers.edu
}

\begin{document}
\maketitle

\begin{abstract}
Graph neural networks (GNNs) are a class of effective deep learning models for node classification tasks; yet their predictive capability may be severely compromised under adversarially designed unnoticeable perturbations to the graph structure and/or node data. Most of the current work on graph adversarial attacks aims at lowering the overall prediction accuracy, but we argue that the resulting abnormal model performance may catch attention easily and invite quick counterattack. Moreover, attacks through modification of existing graph data may be hard to conduct if good security protocols are implemented. In this work, we consider an easier attack harder to be noticed, through adversarially patching the graph with new nodes and edges. The attack is universal: it targets a single node each time and flips its connection to the same set of patch nodes. The attack is unnoticeable: it does not modify the predictions of nodes other than the target. We develop an algorithm, named GUAP, that achieves high attack success rate but meanwhile preserves the prediction accuracy. GUAP is fast to train by employing a sampling strategy. We demonstrate that a 5\% sampling in each epoch yields 20x speedup in training, with only a slight degradation in attack performance. Additionally, we show that the adversarial patch trained with the graph convolutional network transfers well to other GNNs, such as the graph attention network.
\end{abstract}

\section{Introduction}
\label{intro}
Graph structured data are ubiquitous, with examples ranging from molecules, social networks, power systems, to knowledge graphs. Graph representation learning is one of the key areas of machine learning, with several extensively explored downstream tasks including node classification, graph classification, and community detection. Over the past decade, a plethora of learning methods were proposed, ranging from unsupervised embedding approaches (e.g., DeepWalk~\cite{perozzi2014} and node2vec~\cite{grover2016node2vec}) to supervised/semi-supervised graph neural network (GNN) models (e.g., GCN~\cite{Kipf2017} and GAT~\cite{velivckovic2017graph}). GNN models steadily improve the performance of downstream tasks and achieve state-of-the-art results.

The seminal work of \cite{szegedy2013intriguing} and \cite{goodfellow2014explaining} point out that despite achieving high prediction accuracy, deep models are fragile to adversarially manipulated inputs, stirring a proliferation of research on designing adversarial attacks and defense schemes against them. GNNs as an emerging class of deep models tailored for graph structured data also urge scrutiny. A major development in this context focuses on node classifications, in part because of their economic and societal importance. For example, bad actors in a financial network may hide themselves through manipulating contacts and transactions to benign actors, devastating the predictive power of GNNs in identifying illicit activities.

Much prior work~\cite{dai2018adversarial,wu2019adversarial,zugner2019adversarial} studying adversarial attacks on GNNs aims at lowering the classification accuracy toward all nodes in the graph, through either poisoning the training data to weaken training, or modifying the test data to mislead trained models. However, in many practical scenarios, taking control of existing data proves to be challenging and thus modifying the graph data is less realistic.

\begin{figure*}[ht]
  \centering
  \includegraphics[width=\linewidth]{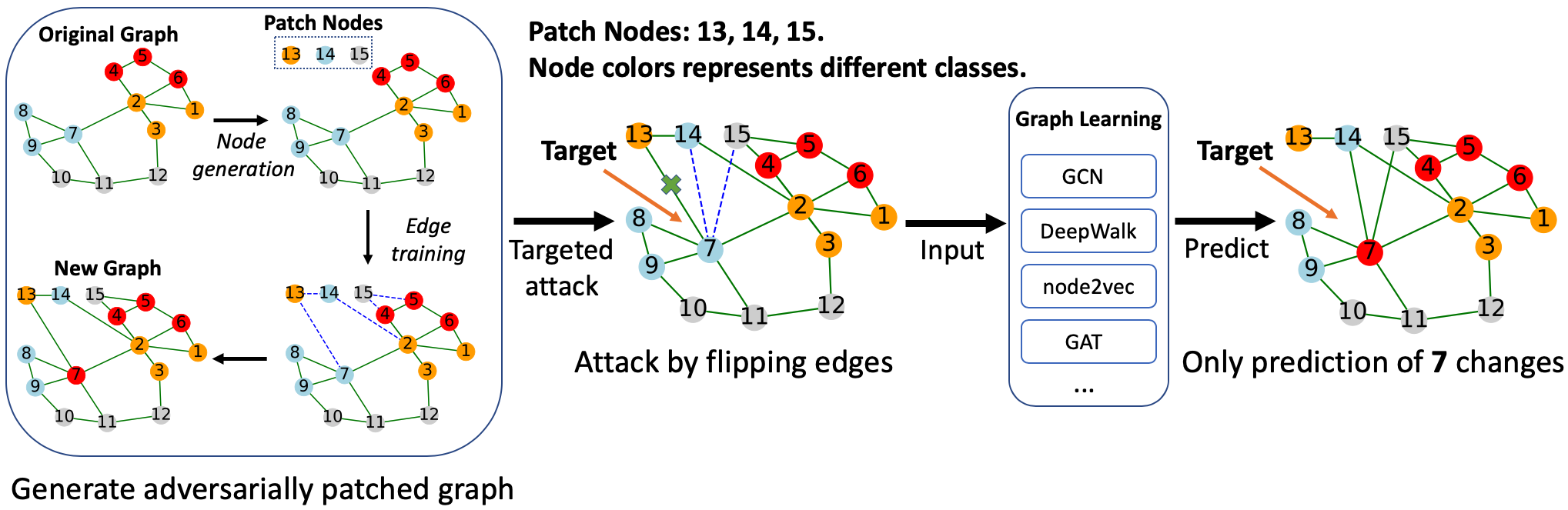}
  \caption{Illustration of GUAP. A set of patch nodes \{13, 14, 15\} and edges are inserted. One attacks node 7 through flipping its connections with the patch. Predictions of other nodes remain unchanged.}
  \label{framework_fg}
\end{figure*}

In this work, we consider attacking a trained model through adversarially patching the graph data with new nodes and edges. These new edges must involve the new nodes; they should not change the connections between existing ones. For example, in a social network setting, the patching amounts to creating new accounts and setting their friendship. The key is that such new nodes are adversarial and their effect is secretive: When a target is being attacked, its connections to the patch nodes are all flipped so that its prediction is changed, while the predictions to other nodes are not. See Figure~\ref{framework_fg} for illustration.

The idea of adversarial patching exists in prior graph-attack work. Greedy-GAN~\cite{wang2018attack} adopts a greedy approach and NIPA~\cite{sun2019node} uses reinforcement learning to compute the new edges. Our work differs from them in several aspects. First, these methods aim at lowering the prediction accuracy whereas ours barely. Hence, they often need a large patch (e.g., 20\% of the original node set in~\cite{wang2018attack} and 10\% in~\cite{sun2019node}, for the Cora data set) but our patch is rather small (e.g., 1\%). Second, these methods attack the entire node set all at once whereas ours targets a single node each time. Third, even though the work of~\cite{wang2018attack} additionally considers attacking single targets, such an attack needs a new optimization whenever the target changes, incurring expensive computation. On the contrary, our work computes the patch once for all and uses the flipping mechanism to perform attack, which is computationally economic. Our attack is of the universal type.

We propose an algorithm, named GUAP, to compute the adversarial patch. It consists of two parts: node generation and edge training. Features of the new nodes are random and they are generated based on the statistics of those of the original graph. We find that the random generation is robust in the sense that once the patch is computed, regenerating the node features using the same mechanism barely affects the attack performance. For edge training, we treat the connections involving the new nodes as parameters to optimize. The optimization achieves two goals: (i) it alters the prediction of the attack target and (ii) it maintains the prediction of other nodes. The latter goal distinguishes our work from most of the prior work.

We summarize the contributions as follows:
\begin{enumerate}[leftmargin=*]
\item We present a novel attack scenario, which patches given graph data without modifying its original content and attacks one target at a time through flipping its connections to the patch nodes.

\item We propose a universal attack algorithm, which achieves high attack success rate (ASR) while maintaining the original prediction accuracy.

\item We show that attack training can be speeded up through sampling the training set in each epoch, without sacrificing the attack performance a lot.

\item We demonstrate that our method admits good transferability of attack performance to a model different from the one used for training.
\end{enumerate}

\section{Related Work}
\paragraph{Universal attack.} Universal attacks compute input-independent perturbations to fool the classifier. They more often appear in the computer vision literature. The work of~\cite{moosavi2017universal} perturbs every pixel of an image whereas the work of~\cite{46561} computes an adversarial patch that is attached to an image at random locations. For graphs, research is sporadic. The work of~\cite{zang2020graph} selects a set of anchor nodes so that attacking a target amounts to flipping the connections between the target and the anchors.

\paragraph{Graph adversarial attack.} Much work devotes to the modification of the graph structure, resulting in poor quality of node embeddings~\cite{chen2018fast,xu2019adversarial,wang2019attacking,bojchevski2018adversarial,dai2018adversarial,xu2019topology}. Nettack~\cite{zugner2018adversarial} modifies not only the graph structure but also the node features. Specifically, targeting each node, Nettack modifies the node feature entry or graph structure step by step, to maximize the prediction loss of the attacked node in a greedy manner. Modifying the graph is generally a discrete optimization problem, which invites greedy algorithms, but the work of~\cite{liu2019unified,bose2019generalizable} proposes a probabilistic framework under which continuous optimization is performed. The work of~\cite{zugner2019adversarial} poisons the graph by treating it as a hyperparameter and optimizing through hypergradient updates. Closest to our work are~\cite{wang2018attack} and~\cite{sun2019node}, both of which inject adversarial nodes to the graph. The former greedily inserts nodes and uses a discriminator to compute node features so that one cannot distinguish new nodes from the original ones. The latter computes adversarial nodes by using reinforcement learning. Neither approach attacks a single node through connection flipping as we do.

\section{Preliminaries}
\label{pre}
We use GCN~\cite{Kipf2017} as the attack model. Denote by $G=(A,X)$ the given graph, where $A$ is the $n\times n$ adjacency matrix and $X$ is the $n\times d$ node feature matrix. Throughout, we assume that the graph is unweighted; thus, $A$ is binary. Denote by $f(A,X)$ the GCN model, which reads
\begin{equation}
Z := f(A,X) = \text{softmax}(\widehat{A}\cdot\relu(\widehat{A}XW^{(0)})\cdot W^{(1)}),
\end{equation}
where $\widehat{A}=\widetilde{D}^{-\frac{1}{2}}\widetilde{A}\widetilde{D}^{-\frac{1}{2}}$ is the normalized adjacency matrix with $\widetilde{A} = A + I$ and $\widetilde{D} = \diag(\sum_j\widetilde{A}_{ij})$. The matrices $W^{(0)}$ and $W^{(1)}$ are trainable parameters whose sizes are respectively $d\times d'$ and $d'\times K$, where $K$ is the number of classes. Consequently, $Z$ is the $n\times K$ output matrix, whose $i$th row is the probability vector for the $i$th node. Let $V_L$ be the training set and $Y$ be the labels. The training of GCN minimizes the cross-entropy loss
\begin{equation}
  L = - \sum_{i\in V_L}\sum_{k=1}^{K}1\{Y_{i} = k\}\ln Z_{ik}.
\end{equation}

\section{Graph Universal Attack Through Adversarial Patching}
Denote by $G_{new}=(A_{new},X_{new})$ the new graph with $m$ patch nodes. For convenience we order them after the original nodes, so that $A_{new} = \begin{bmatrix}A & C \\ C^{T} & B \end{bmatrix}$ and $X_{new} = \begin{bmatrix}X \\ X_{patch} \end{bmatrix}$. Here, $C$ is $n\times m$, denoting the connections between the original nodes and the patch ones; $B$ is $m\times m$, denoting the connections between the patch nodes themselves; and $X_{patch}$ is $m\times d$, denoting the feature matrix of the patch nodes. We will discuss the generation of $X_{patch}$ and the computation of $C$ and $B$ in the following subsections, respectively.

Additionally, we only consider undirected graphs in this paper. This is because even for directed ones, a majority of graph neural networks (e.g., GCN~\cite{Kipf2017}) remove the edge directions and take the symmetric adjacency matrix as input. Our method is evaluated on three undirected graph benchmark data sets.

\subsection{Node Generation}
Realistic node features may be generated by using a generative model (e.g., GAN)~\cite{wang2018attack}, but the learning from existing nodes suffers many challenges, including high dimensionality and small training set. We opt for a simple mechanism that is sufficiently robust.

We treat each feature dimension independently. In general, for numeric features we fit a normal distribution for each feature dimension and sample from it. Without a priori knowledge, a normal distribution appears to be the most straightforward parameterization. Depending on data sets, more accurate distributions may be fit or even learned.

For some of the data sets we experiment with, the node features are binary. Hence, we perform binarization and make the feature value 0 if the Gaussian sample is smaller than 0.5, or 1 otherwise. If the training set contains 1 with probability $p$ and 0 with probability $1-p$, then the fitted normal distribution has mean $p$ and variance $p(1-p)$. Thus, the new samples take 1 with probability $\frac{1}{2}\left[1-\erf\left(\frac{1/2-p}{\sqrt{2p(1-p)}}\right)\right]$, which is approximately $p$. In other words, the general approach of fitting a normal distribution covers well the special case of Bernoulli distribution.

\subsection{Edge Training}
Denote by $\hat{l}(A,X,i)$ the predicted label of node $i$ given graph adjacency matrix $A$ and node feature matrix $X$. It is where the largest entry of $f(A,X)_i$ resides; i.e., $\hat{l}(A, X, i) = \argmax f(A, X)_i$. Our attack aims at two goals: changing the prediction of $i$ while preserving those of other nodes. Both goals involve the graph adjacency matrix $A_{new}'$ when node $i$ is being attacked. They may be mathematically summarized as:
\begin{multline}
  \label{eqn:constraint}
  \text{for each $i$ in the training set $V_L$,} \\ 
  \begin{cases}
  \hat{l}(A_{new}', X_{new}, i) \neq \hat{l}(A, X, i), \\
  \hat{l}(A_{new}', X_{new}, j) = \hat{l}(A, X, j), \quad\forall j\ne i.
  \end{cases}
\end{multline}
Note that $A_{new}'$ is $i$-dependent but we suppress the dependency in notation to avoid cluttering.

We elaborate how $A_{new}'$ is computed from $A_{new}$. Let $p=[0,\ldots,0,1,\ldots,1]$ be an $(n+m)$-vector, where the first $n$ entries are 0 and the rest are 1. We call $p$ the \emph{attack vector}, since the 1 entries will be used to flip the connections with the patch nodes. We extend $p$ to the \emph{attack matrix} $P$, whose $i$th row and $i$th column are the same as $p$ and zero otherwise. Thus, $P_{ij}$ denotes whether the connection between node $i$ and $j$ is flipped. One easily derives that
\begin{equation}
\label{perturb_eq}
  A_{new}' := \text{attack}(A_{new},i) = (\mathbbm{1} - P) \circ A_{new} + P \circ (\mathbbm{1}_0 - A_{new}) ,
\end{equation}
where $\circ$ stands for element-wise product, $\mathbbm{1}$ is a matrix of all ones, and $\mathbbm{1}_0$ is analogous except that the diagonal is set as zero.

Throughout training, we will also need to revert the attacked graph back to the patched graph. Such an ``unattack'' operation is simple to conduct by using the attack matrix to flip back:
\begin{equation}
  A_{new} := \text{unattack}(A_{new}',i) = \text{attack}(A_{new}',i).
\end{equation}

\subsubsection{Outer Loop: GUAP}
Recall that $A_{new} = \begin{bmatrix}A & C \\ C^{T} & B \end{bmatrix}$. The overall algorithm is to start with an initial $A_{new}$ (specifically, $B=0$ and $C=0$) and iteratively update it by using certain perturbation $\Delta A_{new} = \begin{bmatrix}0 & \Delta C \\ \Delta C^{T} & \Delta B \end{bmatrix}$ that reflects the two goals summarized in~\eqref{eqn:constraint}. See Algorithm~\ref{gpa_alg}.

\begin{algorithm}[h]
  \caption{Graph Universal Attack Through Adversarial Patching (GUAP)}
  \label{gpa_alg}
  \begin{algorithmic}
    \STATE \textbf{Input:} $A$, $X$
    \STATE \textbf{Output:} Adjacency matrix $A_{new}$ of the patched graph and node features $X_{new}$
    \STATE Initialize $A_{new}$ and generate $X_{new}$
    \STATE $epoch \leftarrow 0$
    \WHILE{ $epoch < max\_epoch$}
    \FOR{node $i$ in training set}
    \STATE $A_{new}' \leftarrow \text{attack}(A_{new},i)$
    \IF{$\hat{l}(A_{new}', X_{new}, i) = \hat{l}(A, X, i)$}
    \STATE $\Delta A_{new} \leftarrow \text{IGP}(A_{new}', X_{new}, i)$
    \STATE $A_{new}' \leftarrow A_{new}' + \Delta A_{new}$
    \STATE $A_{new} \leftarrow \text{unattack($A_{new}'$, $i$)}$
    \STATE $A_{new} \leftarrow \text{$L_2$-projection}(A_{new},radius)$
    \STATE $A_{new} \leftarrow A_{new}.clip(0,1)$
    \STATE $A_{new}.diagonal \leftarrow 0$
    \ENDIF
    \ENDFOR
    \STATE $A_{new} \leftarrow (A_{new}>0.5)\,?\,1:0$
    \STATE Compute ASR using Equation~\eqref{asr_eq} and record the highest value
    \STATE $epoch \leftarrow epoch + 1$
    \ENDWHILE
    \STATE \textbf{return} $A_{new}$ at the epoch of highest ASR and $X_{new}$
  \end{algorithmic}
\end{algorithm}

Concretely, the training is conducted in several epochs, each of which iterates over the training set $V_L$. At node $i$, we compute the attacked adjacency matrix $A_{new}'$ and check if $i$'s prediction changes. If not, we use an inner procedure IGP (to be elaborated subsequently) to generate a perturbation $\Delta A_{new}$. Then we update $A_{new}'$ with this perturbation and revert it to the unattacked matrix $A_{new}$. Because the perturbation may gradually modify $A_{new}$ to an incomparable magnitude, we apply $L_2$ projection as well as clipping to prevent $A_{new}$ from exploding. The $L_2$ projection is applied to each patch node indivually so that the vector of edges to such a node has an $L_2$ norm $radius$. We also set the diagonal of $B$ to be zero to prevent self loops.

After the entire training set is iterated, the $B$ and $C$ blocks contain real values within $(0,1)$. We binarize them to maintain unweightedness. Then, the attack success rate
\begin{equation}
\label{asr_eq}
  ASR(V_{L}) := \frac{1}{|V_L|} \sum\limits_{i=1}^{|V_L|}1\{\hat{l}(A_{new}', X_{new}, i) \neq \hat{l}(A, X, i)\}
\end{equation}
is computed as the metric of attack performance.

\subsubsection{Inner Loop: IGP}
The inner procedure \emph{iterative graph perturbation}, IGP, computes a perturbation $\Delta A_{new}$ to the current attacked matrix $A_{new}'$ to gear it toward the two goals summarized in~\eqref{eqn:constraint}. For the first goal, the strategy is to push the prediction toward the decision boundary of another class; whereas for the second goal, the strategy is to progress toward a smaller loss for all nodes except $i$:
\begin{equation}
\label{newloss_eq}
L_{new}' := - \sum_{j\in V_L \setminus i}\,\sum_{k=1}^{K}1\{Y_{j} = k\}\ln f(A_{new}', X_{new})_{jk}.
\end{equation}
The procedure is summarized in Algorithm~\ref{graphfool_alg}.

\begin{algorithm}[ht]
  \caption{Iterative Graph Perturbation (IGP)}
  \label{graphfool_alg}
  \begin{algorithmic}
    \STATE \textbf{Input:} Attacked adjacency matrix $A_{new}'$, feature matrix $X_{new}$, node $i$
    \STATE \textbf{Output:} Perturbation $\Delta A_{new}$
    \STATE Initialize empty $\Delta A_{new}$
    \STATE $E_{new}' \leftarrow A_{new}'$
    \STATE $iter \leftarrow 0$
    \STATE $pred \leftarrow \hat{l}(A, X, i)$
    \WHILE{$\hat{l}(A_{new}', X_{new}, i) = pred$ \text{and} $iter < max\_iter$}
    \STATE $v \leftarrow \frac{|\Delta f_{k}|}{||\Delta w_{k}||_2^2} \Delta w_{k} \text{ according to Equation~\eqref{deepfool_eq}}$
    \STATE $v[0:n] \leftarrow 0$
    \STATE $\Delta A_{new}[i,:] \leftarrow \Delta A_{new}[i,:] + (1+overshoot) \cdot v$ and analogously for $\Delta A_{new}[:,i]$
    \STATE $E_{new}' \leftarrow A_{new}' + \Delta A_{new}$
    \STATE $E_{new}' \leftarrow E_{new}'.clip(0, 1)$
    \STATE $grad \leftarrow \nabla L_{new}'(E_{new}')$ \quad\# see loss function~\eqref{newloss_eq}
    \STATE $grad[0:n, 0:n] \leftarrow 0$
    \STATE $grad[i,:] \leftarrow 0$ and analogously for $grad[:,i]$
    \STATE $grad \leftarrow (grad + grad^T) / 2$
    \STATE $\Delta A_{new} \leftarrow \Delta A_{new} - step \cdot  grad$
    \STATE $iter \leftarrow iter + 1$
    \ENDWHILE
    \STATE \textbf{return} $\Delta A_{new}$
  \end{algorithmic}
\end{algorithm}

Algorithm~\ref{graphfool_alg} is a while-loop that iteratively computes the perturbation till the prediction of node $i$ changes (or the iteration count reaches maximum). The loop contains two part, corresponding to the two goals respectively. The first part intends to attack $i$. Denote the prediction $pred$; i.e., $pred = \hat{l}(A, X, i)$. According to~\cite{moosavi2016deepfool,zang2020graph}, the minimum perturbation $v$ on the $i$th row (and column) of $A_{new}'$ that sends node $i$ to the decision boundary of the closest class $k$ can be calculated as
\begin{equation}
\label{deepfool_eq}
    k = \argmin_{c \neq pred} \frac{|\Delta f_c|}{\|\Delta w_c\|_2},\,\,\,
    v = \frac{|\Delta f_{k}|}{||\Delta w_{k}||_2^2} \Delta w_{k} ,
\end{equation}
where $\Delta f_c = f(A_{new}, X_{new})_{i,c} - f(A_{new}, X_{new})_{i,pred}$ and $\Delta w_c = \nabla f(A_{new}, X_{new})_{i,c} - \nabla f(A_{new}, X_{new})_{i, pred}$.
Here, gradient is taken with respect to the $i$th row (and symmetrically column) of $A_{new}$. We set the first $n$ entries of $v$ to be zero because the original graph should not change. We also use a small $overshoot$ constant to send node $i$ to the other side of the decision boundary. We introduce a temporary notation $E_{new}'$ to denote the updated $A_{new}'$ for subsequent use. We also apply clipping, similar to what is done in the outer loop.

The second part intends to lower the loss~\eqref{newloss_eq}. We calculate its gradient $grad$ at $E_{new}'$ and set the first $n\times n$ block to be zero, We also set the $i$th row and column zero because prediction of node $i$ is not to be preserved. After numerical symmetrization, we update the perturbation $\Delta A_{new}$ along the gradient descent direction, completing one iteration of the while-loop.

\section{Experiments}
In this section, we perform a set of comprehensive experiments to demonstrate the effectiveness of GUAP. We investigate the patch size, show speedup of training under the sampling strategy, compare with several related methods, and present transferability results. Code is available at \url{https://anonymous.4open.science/r/ffd4fad9-367f-4a2a-bc65-1a7fe23d9d7f/}.

\subsection{Data Sets and Details}
We use three commonly used benchmark data sets: Cora, Citeseer, and Pol.Blogs.
Their information is summarized in Table~\ref{dataset_tb}. Training is based on the standard GCN model and hence we also list its test accuracy in the table.

\begin{table}[ht]
  \centering
  \caption {Node Classification Datasets. Only the largest connected component (LCC) is considered.}
  \label{dataset_tb}
  \begin{tabular}{lccccc}
    \toprule
    Statistics & Cora & Citeseer & Pol.Blogs \\ 
    \midrule
    \# Nodes(LCC) & $2708$ & $3327$ & $1222$ \\
    \# Edges(LCC) & $5278$ & $4676$ & $16714$ \\
    \# Classes & $7$ & $6$ & $2$ \\
    Train/teset Set & $140/1000$ & $120/1000$ & $121/1101$ \\
    Accuracy(GCN) & $81.4\%$ & $70.4\%$ & $94.3\%$ \\
    \bottomrule
  \end{tabular}
\end{table}

The hyperparameters for Algorithms~\ref{gpa_alg} and~\ref{graphfool_alg} are: $max\_epoch = 50$, $max\_iter = 30$, $radius = 10$, $overshoot=0.02$, and $step = 10$. All experiments are repeated ten times under the same hyperparameters.

\begin{figure*}[h]
  \centering
  \subfigure[Cora.]{
    \includegraphics[height=1.5in]{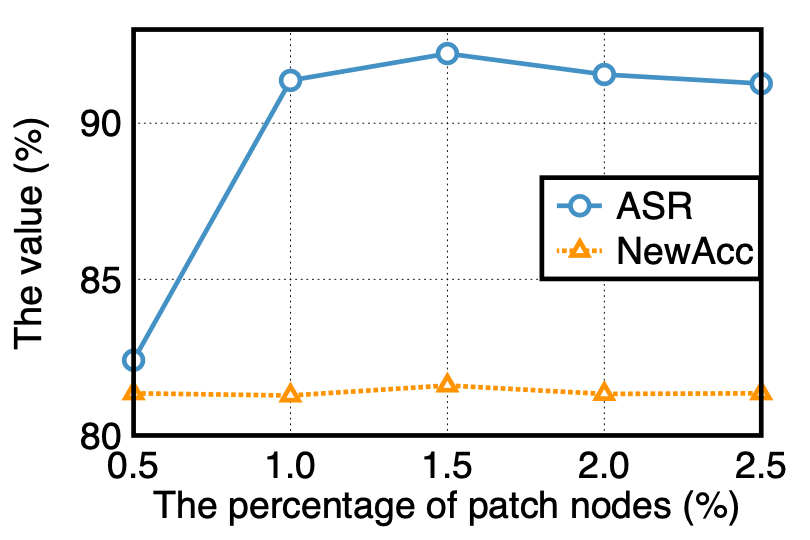}}
  \subfigure[Citeseer.]{
    \includegraphics[height=1.5in]{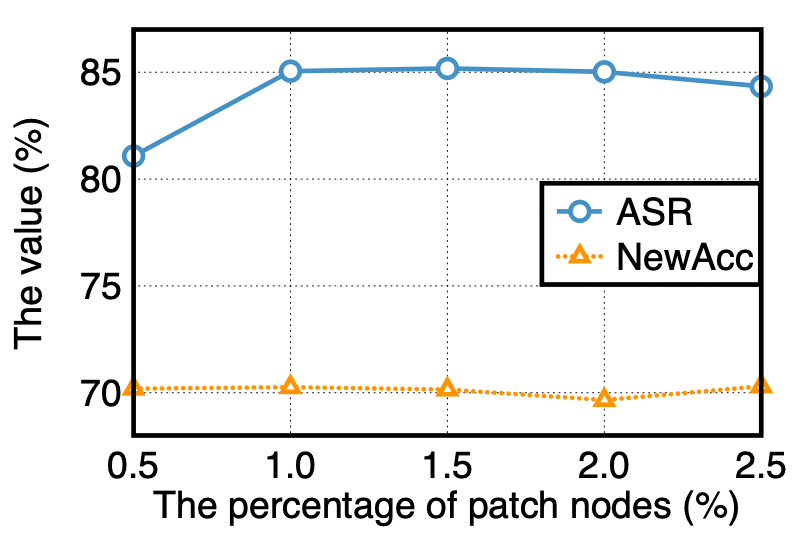}}
  \subfigure[Pol.Blogs.]{
    \includegraphics[height=1.5in]{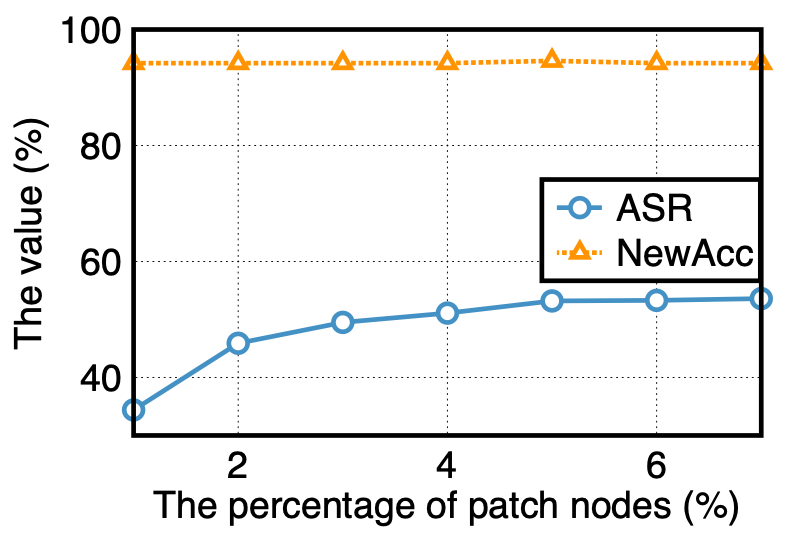}}
  \caption{ASR and accuracy as number of patch nodes increases (in percentage of the node set size).}
  \label{patchsize_fg}
\end{figure*}

\begin{figure*}[h]
  \centering
  \subfigure[Cora.]{
    \includegraphics[height=1.5in]{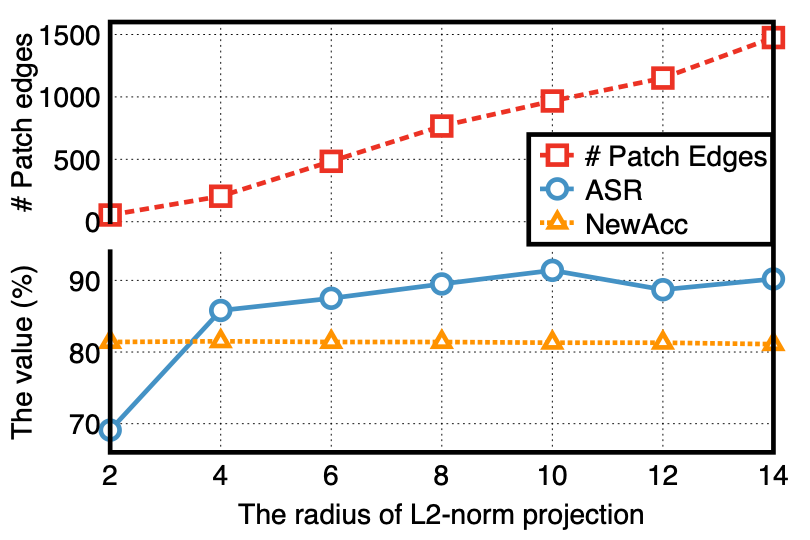}}
  \subfigure[Citeseer.]{
    \includegraphics[height=1.5in]{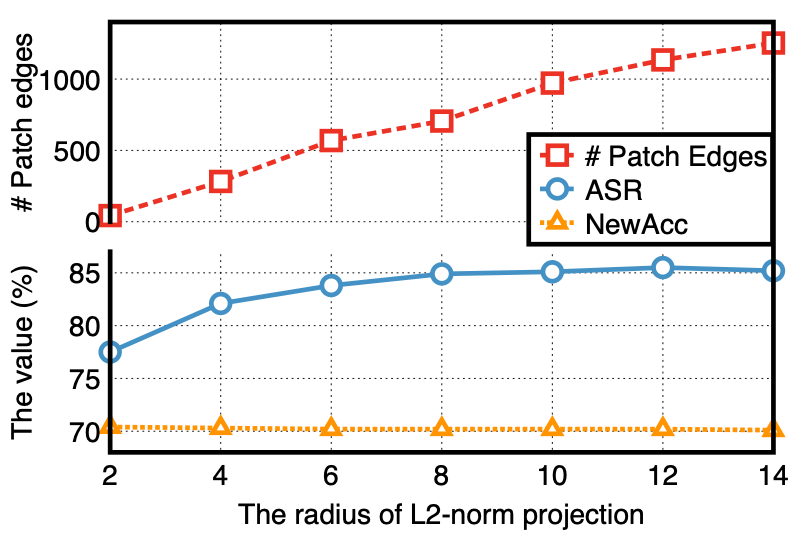}}
  \subfigure[Pol.Blogs.]{
    \includegraphics[height=1.5in]{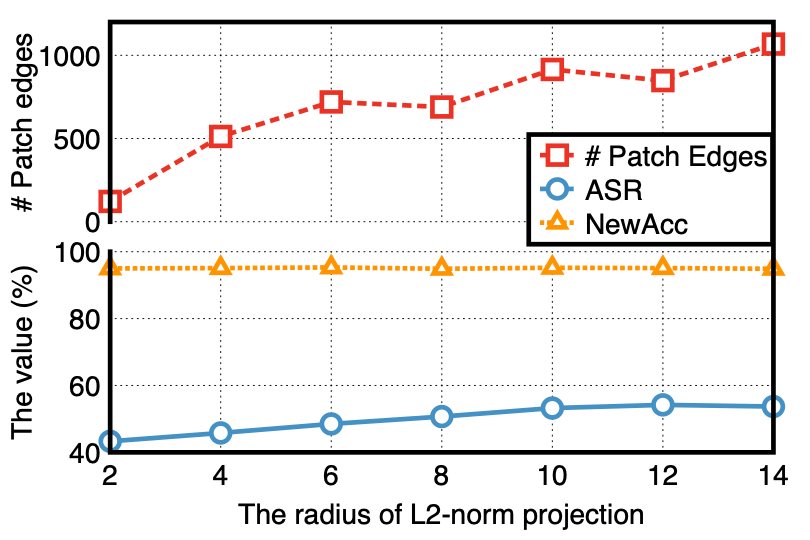}}
  \caption{Average number of patch edges, ASR, and accuracy as the $L_2$ projection radius $\xi$ increases.}
  \label{patchedge_fg}
\end{figure*}
\subsection{Compared Methods}
\label{baseline_sec}
Universal attacks on graphs are rarely studied; hence, for comparison are a combination of existing universal attack method, non-universal method, and variants of the proposed method.

\begin{enumerate}[leftmargin=*]
\item Graph Universal Attack (GUA)~\cite{zang2020graph}. As opposed to adversarial patching, GUA seeks a set of anchor nodes from the graph and attacks a target through flipping its connections to these anchors. For a fair comparison, the number of anchors is the same as the patch size in GUAP.

\item Fast Gradient Attack (FGA)~\cite{chen2018fast}. FGA is not a universal attack method. Targeting each node, it iteratively modifies the patch connection with the largest absolute gradient value. For a fair comparison, FGA can modify up to the same number of patch edges as GUAP. 

\item GUAP without patching edges. This variant of GUAP introduces only patch nodes but no edges. In other words, when a node is attacked, it will be connected through edges to all patch nodes.

\item GUAP with randomly patched edges. Rather than performing the sophisticated edge training, this variant introduces random edges to the patch nodes. In this case, the existence of an edge follows a Bernoulli distribution with certain success probability. We experiment with two cases: one such that the number of new edges is approximately the same as the that in GUAP; and the other merely setting probability $=0.5$, introducing many more edges.

\item GUAP with regenerated node features. This variant first computes the patched graph as does GUAP; then, it regenerates the patch node features. Note that the training of the patched graph relies on the initial features. Hence, it is interesting to see how change of features affects attack.
\end{enumerate}

\subsection{Results}
We use two metrics for evaluation: ASR and $\Delta$Acc (change of prediction accuracy).

\paragraph{Patch size.}
First we determine a reasonable patch size. Figure~\ref{patchsize_fg} reveals a common pattern across data sets: the ASR increases as more and more nodes are patched into the graph, before peak, whereas the accuracy is fairly stable. The ASR curve climbs quickly, indicating that a small patch size suffices to achieve good ASR. We thus set the patch size to be 1\%, 1\%, and 5\% of the original node set for Cora, Citeseer, and Pol.Blogs, respectively.

\begin{table}
\centering
\caption{Average ASR of GUAP with and without clipping. The percentages of the patch nodes are $1\%$, $1\%$, $5\%$ for Cora, Citeseer, and Pol.Blogs, respectively. The projection radius $\xi = 10$. }
\label{noclip_tb}
\begin{tabular}{l|ccc}
\toprule
Method & Cora & Citeseer & Pol.Blogs \\
\midrule
GUAP w/o clipping & $82.24\%$ & $83.41\%$ & $45.76\%$ \\
GUAP w/ clipping & $91.37\%$ & $85.05\%$ & $53.24\%$ \\
\bottomrule
\end{tabular}

\end{table}

\paragraph{The $L_2$ Projection Radius.}
We then keep the plateaued percentages of patch nodes found in Figure~\ref{patchsize_fg} and investigate the influence of one important hyper-parameter: $L_2$ projection radius $\xi$. In Figure~\ref{patchedge_fg}, we report the average ASR, prediction accuracy, and the number of patch edges by increasing $\xi$. It shows that when $\xi = 10$, the ASRs achieve the highest value on the three benchmarks, while preserving the overall prediction accuracy. Afterward, the ASR will not increase further. 

Moreover, the number of patch edges quickly climbs up with larger $\xi$. This is because the number of patch edges is implicitly controlled by the projection radius. Increasing $\xi$ will densify the patch. In real situations, we can adjust the projection radius to make a balance in the trade-off between ASR and number of patch edges, so that the edge density for the added patches is indistinguishable from real ones. Nevertheless, in subsequent experiments, we adopt $\xi=10$ for the highest ASR, regardless of the density of patch edges.

\paragraph{Necessity of clipping.}
Based on~\cite{zang2020graph}, we also adopt clipping to encourage the stability of results. In Table~\ref{noclip_tb}, we list the average ASR of two variants of GUAP using clipping versus not. It shows that clipping significantly increases the ASR of GUAP. Therefore, clipping is a necessary ingredient of the method for achieving high attack performance.

\paragraph{Training cost and acceleration.}
Next we investigate the computational cost. An estimate is $O(max\_epoch \cdot |V_L| \cdot m(m+2n))$, where the factor $|V_L|$ (denoting the training set size) comes from the for-loop in Algorithm~\ref{gpa_alg}, whereas the factor $m(m+2n)$ (denoting the difference in matrix size between the original graph and the patched graph) comes from the inner procedure Algorithm~\ref{graphfool_alg}. Because the node set size $n$ is given and the patch size $m$ is implicitly controlled by the desired ASR (see the preceding experiment), one factor we may adjust to scale the cost better is the length of the for-loop. Inside each epoch, rather than iterating over the entire training set, we deal with a random subset only. Table~\ref{sample_tb} shows that as one uses a smaller subset, the training time reduces proportionally, whereas ASR suffers only slightly and accuracy barely changes. Hence, for a large graph with large training set, the sampling scheme effectively accelerates training.

\begin{table*}[ht]
  \caption{ASR and change of accuracy under different sampling rate of the training set. Values in () next to the data set name denote the patch set size as a percentage of the node set size.}
  \label{sample_tb}
  \centering
  \begin{tabular}{lcccccc}
    \toprule
    & \multicolumn{2}{c}{Cora (1\%)}&\multicolumn{2}{c}{Citeseer (1\%)}&\multicolumn{2}{c}{Pol.Blogs (5\%)} \\
    \cmidrule(r){2-3}
    \cmidrule(r){4-5}
    \cmidrule(r){6-7}
    Sample Rate & ASR & $\Delta$Acc & ASR & $\Delta$Acc & ASR & $\Delta$Acc \\
    \midrule
    100\%               & $91.37\%$ & $-0.12\%$ & $85.05\%$ & $-0.14\%$ & $53.24\%$ & $+0.26\%$ \\
    40\% (3x speedup) & $89.57\%$ & $-0.17\%$ & $86.93\%$ & $-0.14\%$ & $52.92\%$ & $+0.35\%$ \\
    20\% (5x speedup)   & $87.25\%$ & $-0.17\%$ & $87.53\%$ & $-0.19\%$ & $53.01\%$ & $+0.26\%$ \\
    10\% (10x speedup)   & $82.42\%$ & $+0.01\%$ & $83.41\%$ & $-0.15\%$ & $52.78\%$ & $+0.36\%$ \\
    5\% (20x speedup)   & $80.01\%$ & $-0.16\%$ & $79.43\%$ & $-0.09\%$ & $52.77\%$ & $+0.36\%$ \\
    \bottomrule
  \end{tabular}
\end{table*}

\begin{table*}[h]
  \caption{Comparison of with related methods. Values in () next to the data set name denote the patch set size.}
  \label{comp_tb}
  \centering
  \begin{tabular}{lcccccc}
    \toprule
    & \multicolumn{2}{c}{Cora (29)}&\multicolumn{2}{c}{Citeseer (33)}&\multicolumn{2}{c}{Pol.Blogs (45)} \\
    \cmidrule(r){2-3}
    \cmidrule(r){4-5}
    \cmidrule(r){6-7}
    Baselines & ASR & $\Delta$Acc & ASR & $\Delta$Acc & ASR & $\Delta$Acc \\
    \midrule
    GUA                    & $86.48\%$ & $-0.07\%$ & $82.23\%$ & $-0.07\%$ & $48.36\%$ & $+0.38\%$ \\
    GUAP w/o patch edges   & $28.34\%$ & $-0.01\%$ & $25.02\%$ & $-0.01\%$ & $14.62\%$ & $+0.39\%$ \\
    GUAP w/ random edges  &  $58.27\%$ & $-0.79\%$ & $62.73\%$ & $-0.88\%$ & $19.99\%$ & $+0.31\%$ \\
    GUAP w/ more rand. edges   & $68.81\%$ & $-46.03\%$& $77.74\%$ &$-48.47\%$ & $18.69\%$ & $-17.26\%$ \\
    GUAP                   & $91.42\%$ & $-0.11\%$ & $85.03\%$ & $-0.15\%$ & $51.10\%$ & $+0.36\%$ \\
    GUAP + regen. features & $91.41\%$ & $-0.02\%$ & $85.00\%$ & $-0.02\%$ & $51.08\%$ & $+0.36 \%$ \\
    FGA (Not universal)    & $94.90\%$ & $-0.66\%$ & $92.91\%$ & $-0.20\%$ &$42.74\%$ & $-0.14\%$ \\
    \bottomrule
  \end{tabular}
\end{table*}


\begin{table}[h]
  \caption{Attack performance when using the patched graph trained with GCN on other models. The patch set percentage is $1\%$, $1\%$, $5\%$ for Cora, Citeseer and Pol.Blogs, respectively.}
  \label{transf_tb}
  \centering
  \begin{tabular}{l|ccc}
    \toprule
    Methods & Cora & Citeseer & Pol.Blogs \\
    \midrule
    GCN(ASR)      & $91.37\%$  & $85.05\%$ & $53.24\%$ \\
    GCN($\Delta$Acc)        & $-0.12\%$ & $-0.14\%$ & $+0.26\%$ \\
    \midrule
    GAT(ASR) & $90.91\%$ & $85.04\%$ & $40.02\%$ \\
    GAT($\Delta$Acc) & $-0.36\%$ & $-0.19\%$ & $-0.04\%$ \\
    \midrule
    node2vec(ASR) & $74.89\%$ & $84.24\%$ & $43.07\%$ \\
    node2vec($\Delta$Acc) & $+2.58\%$ & $+3.66\%$ & $+2.83\%$ \\
    \midrule
    DeepWalk(ASR) & $81.02\%$ & $82.41\%$ & $41.32\%$ \\
    DeepWalk($\Delta$Acc) & $-41.42\%$ & $-21.51\%$ & $-3.50\%$ \\
    \midrule
    FastGCN(ASR) & $41.39\%$ & $34.74\%$ & $36.59\%$ \\
    FastGCN($\Delta$Acc) & $-2.43\%$ & $-0.40\%$ & $-2.08\%$ \\
    \midrule
    AS-GCN(ASR) & $36.68\%$ & $31.09\%$ & $39.24\%$ \\
    AS-GCN($\Delta$Acc) & $-2.42\%$ & $-1.46\%$ & $-2.24\%$ \\
    \bottomrule
  \end{tabular}
\end{table}

\paragraph{Comparison with related methods.}
Now we compare GUAP with several of its variants, as well as GUA and FGA. See Table~\ref{comp_tb}. Same as GUAP, GUA preserves the prediction accuracy but achieves a lower ASR. The preservation of accuracy indicates that 
most nodes have a robust neighborhood and the compromise of only the target as one of the neighbors affects little. The observation similarly applies to GUAP without patching any edges, although in this case the ASR is significantly dropped. The observation also applies to GUAP with randomly patched edges, because the number of such edges is quite small. In this case, the ASR also suffers, although to a less extent than the case of not patching any edges. However, when more and more random edges are patched, these edges play an increasingly significant role to the neighborhood, leading to substantial compromise in the prediction accuracy. Next, GUAP with regenerated node features barely changes the ASR and the accuracy. This observation, together with earlier ones, indicates that node features are much less important than edges, the effort of training which pays. The non-universal attack method FGA also barely changes the accuracy, but in some cases it achieves a higher ASR than GUAP while in others not.

Nettack~\cite{zugner2018adversarial} is a popular non-universal attack method. Due to its high computational cost for a comparable perturbation, it is infeasible to conduct experiments with Nettack in a fair setting. Here, we highlight the computational costs. GUAP takes time $O(max\_epoch \cdot |V_L| \cdot m(m+2n))$. $|V_L|$ is typically a small fraction of $n$ and grows much slower than $n$, which can be further reduced by sampling. On the other hand, to attack all nodes, Nettack costs $O(n^2(E \cdot T+F)))$, where $E$ and $F$ represent the number of edge and feature perturbations, respectively, and $T$ is the average size of a 2-hop neighborhood. In practice, the $n^2$ factor renders Nettack a rather slower method to run, if the aim is to attack all nodes. 


\paragraph{Transferability to other models.}
We apply the patch trained with GCN on other GNN models: GAT~\cite{velivckovic2017graph}, FastGCN~\cite{chen2018fastgcn}, AS-GCN~\cite{huang2018adaptive}, and two embedding models node2vec~\cite{grover2016node2vec} and DeepWalk~\cite{perozzi2014}. Table~\ref{transf_tb} summarizes the results. GAT is developed based on GCN through incorporating the attention mechanism. Node2vec and DeepWalk update node embeddings by exploring the local structure via random walk. Different from the other models, instead of using the whole graph, FastGCN and AS-GCN use importance sampling to sample layer-wise nodes to reduce training cost. 

One sees that the attack performance is well maintained on GAT, except the ASR of Pol.Blogs. For this exception and all cases of node2vec and DeepWalk, the ASR still is reasonably similar to the GCN case. However, the node2vec accuracy surprisingly increases and the DeepWalk accuracy significantly drops. 

Additionally, both FastGCN and AS-GCN reveal robustness against our attack, which is by and large owing to the use of sampling. Such an observation is not surprising. The patch is quite small, constituting only $1\%$ of the nodes in Cora and Citeseer. Consequently, the patch nodes are likely to be ignored in sampling and thus voiding attacks. Furthermore, the patches do not negatively impact the overall accuracy significantly.

Based on the above findings, we see that the patch optimized for GCN is not guaranteed to work similarly on all other models, although it does perform equally well on GAT and also reasonably close on DeepWalk and node2vec in terms of ASR. Such a result is expected, since GAT has a similar architecture to GCN whereas the other models operate quite differently. Overall, we conclude that our approach transfers well to neural architectures similar to the one trained on.

\section{Conclusion}
In this paper, we consider a novel type of graph universal attacks that do not modify the existing nodes and edges and that do not change the prediction of nodes other than the target. The attack adversarially patches a small number of new nodes and edges to the original graph. It compromises any target through flipping its connections to the patch. We develop an algorithm, GUAP, to find such a patch and demonstrate high attack success rate. We show that the algorithm can be accelerated through sampling the training set in each epoch without sacrificing attack performance, hinting feasibility for large graphs. For example, a 5\% sampling leads to a 20x speedup in training. GUAP achieves a higher ASR than the recently proposed universal attack GUA. Moreover, the patch trained with GCN can be used to effectively attack other models, such as GAT, as well.

\bibliographystyle{named}
\bibliography{main}

\begin{thebibliography}{}

\bibitem[\protect\citeauthoryear{Bojchevski and
  G{\"u}nnemann}{2018}]{bojchevski2018adversarial}
Aleksandar Bojchevski and Stephan G{\"u}nnemann.
\newblock Adversarial attacks on node embeddings via graph poisoning.
\newblock {\em arXiv preprint arXiv:1809.01093}, 2018.

\bibitem[\protect\citeauthoryear{Bose \bgroup \em et al.\egroup
  }{2019}]{bose2019generalizable}
Avishek~Joey Bose, Andre Cianflone, and William Hamiltion.
\newblock Generalizable adversarial attacks using generative models.
\newblock {\em arXiv preprint arXiv:1905.10864}, 2019.

\bibitem[\protect\citeauthoryear{Brown \bgroup \em et al.\egroup
  }{2017}]{46561}
Tom Brown, Dandelion Mane, Aurko Roy, Martin Abadi, and Justin Gilmer.
\newblock Adversarial patch.
\newblock 2017.

\bibitem[\protect\citeauthoryear{Chen \bgroup \em et al.\egroup
  }{2018a}]{chen2018fastgcn}
Jie Chen, Tengfei Ma, and Cao Xiao.
\newblock Fastgcn: fast learning with graph convolutional networks via
  importance sampling.
\newblock {\em arXiv preprint arXiv:1801.10247}, 2018.

\bibitem[\protect\citeauthoryear{Chen \bgroup \em et al.\egroup
  }{2018b}]{chen2018fast}
Jinyin Chen, Yangyang Wu, Xuanheng Xu, Yixian Chen, Haibin Zheng, and Qi~Xuan.
\newblock Fast gradient attack on network embedding.
\newblock {\em arXiv preprint arXiv:1809.02797}, 2018.

\bibitem[\protect\citeauthoryear{Dai \bgroup \em et al.\egroup
  }{2018}]{dai2018adversarial}
Hanjun Dai, Hui Li, Tian Tian, Xin Huang, Lin Wang, Jun Zhu, and Le~Song.
\newblock Adversarial attack on graph structured data.
\newblock {\em arXiv preprint arXiv:1806.02371}, 2018.

\bibitem[\protect\citeauthoryear{Goodfellow \bgroup \em et al.\egroup
  }{2014}]{goodfellow2014explaining}
Ian~J Goodfellow, Jonathon Shlens, and Christian Szegedy.
\newblock Explaining and harnessing adversarial examples.
\newblock {\em arXiv preprint arXiv:1412.6572}, 2014.

\bibitem[\protect\citeauthoryear{Grover and
  Leskovec}{2016}]{grover2016node2vec}
Aditya Grover and Jure Leskovec.
\newblock node2vec: Scalable feature learning for networks.
\newblock In {\em Proceedings of the 22nd ACM SIGKDD international conference
  on Knowledge discovery and data mining}, pages 855--864. ACM, 2016.

\bibitem[\protect\citeauthoryear{Huang \bgroup \em et al.\egroup
  }{2018}]{huang2018adaptive}
Wenbing Huang, Tong Zhang, Yu~Rong, and Junzhou Huang.
\newblock Adaptive sampling towards fast graph representation learning.
\newblock In {\em Advances in neural information processing systems}, pages
  4558--4567, 2018.

\bibitem[\protect\citeauthoryear{Kipf and Welling}{2017}]{Kipf2017}
Thomas~N. Kipf and Max Welling.
\newblock Semi-supervised classification with graph convolutional networks.
\newblock In {\em ICLR}, 2017.

\bibitem[\protect\citeauthoryear{Liu \bgroup \em et al.\egroup
  }{2019}]{liu2019unified}
Xuanqing Liu, Si~Si, Xiaojin Zhu, Yang Li, and Cho-Jui Hsieh.
\newblock A unified framework for data poisoning attack to graph-based
  semi-supervised learning.
\newblock {\em arXiv preprint arXiv:1910.14147}, 2019.

\bibitem[\protect\citeauthoryear{Moosavi-Dezfooli \bgroup \em et al.\egroup
  }{2016}]{moosavi2016deepfool}
Seyed-Mohsen Moosavi-Dezfooli, Alhussein Fawzi, and Pascal Frossard.
\newblock Deepfool: a simple and accurate method to fool deep neural networks.
\newblock In {\em Proceedings of the IEEE conference on computer vision and
  pattern recognition}, pages 2574--2582, 2016.

\bibitem[\protect\citeauthoryear{Moosavi-Dezfooli \bgroup \em et al.\egroup
  }{2017}]{moosavi2017universal}
Seyed-Mohsen Moosavi-Dezfooli, Alhussein Fawzi, Omar Fawzi, and Pascal
  Frossard.
\newblock Universal adversarial perturbations.
\newblock In {\em Proceedings of the IEEE conference on computer vision and
  pattern recognition}, pages 1765--1773, 2017.

\bibitem[\protect\citeauthoryear{Perozzi \bgroup \em et al.\egroup
  }{2014}]{perozzi2014}
Bryan Perozzi, Rami Al-Rfou, and Steven Skiena.
\newblock Deepwalk: Online learning of social representations.
\newblock In {\em Proceedings of the 20th ACM SIGKDD international conference
  on Knowledge discovery and data mining}, pages 701--710. ACM, 2014.

\bibitem[\protect\citeauthoryear{Sun \bgroup \em et al.\egroup
  }{2019}]{sun2019node}
Yiwei Sun, Suhang Wang, Xianfeng Tang, Tsung-Yu Hsieh, and Vasant Honavar.
\newblock Node injection attacks on graphs via reinforcement learning.
\newblock {\em arXiv preprint arXiv:1909.06543}, 2019.

\bibitem[\protect\citeauthoryear{Szegedy \bgroup \em et al.\egroup
  }{2013}]{szegedy2013intriguing}
Christian Szegedy, Wojciech Zaremba, Ilya Sutskever, Joan Bruna, Dumitru Erhan,
  Ian Goodfellow, and Rob Fergus.
\newblock Intriguing properties of neural networks.
\newblock {\em arXiv preprint arXiv:1312.6199}, 2013.

\bibitem[\protect\citeauthoryear{Veli{\v{c}}kovi{\'c} \bgroup \em et al.\egroup
  }{2017}]{velivckovic2017graph}
Petar Veli{\v{c}}kovi{\'c}, Guillem Cucurull, Arantxa Casanova, Adriana Romero,
  Pietro Lio, and Yoshua Bengio.
\newblock Graph attention networks.
\newblock {\em arXiv preprint arXiv:1710.10903}, 2017.

\bibitem[\protect\citeauthoryear{Wang and Gong}{2019}]{wang2019attacking}
Binghui Wang and Neil~Zhenqiang Gong.
\newblock Attacking graph-based classification via manipulating the graph
  structure.
\newblock In {\em Proceedings of the 2019 ACM SIGSAC Conference on Computer and
  Communications Security}, pages 2023--2040, 2019.

\bibitem[\protect\citeauthoryear{Wang \bgroup \em et al.\egroup
  }{2018}]{wang2018attack}
Xiaoyun Wang, Joe Eaton, Cho-Jui Hsieh, and Felix Wu.
\newblock Attack graph convolutional networks by adding fake nodes.
\newblock {\em arXiv preprint arXiv:1810.10751}, 2018.

\bibitem[\protect\citeauthoryear{Wu \bgroup \em et al.\egroup
  }{2019}]{wu2019adversarial}
Huijun Wu, Chen Wang, Yuriy Tyshetskiy, Andrew Docherty, Kai Lu, and Liming
  Zhu.
\newblock Adversarial examples for graph data: Deep insights into attack and
  defense.
\newblock In {\em International Joint Conference on Artificial Intelligence,
  IJCAI}, pages 4816--4823, 2019.

\bibitem[\protect\citeauthoryear{Xu \bgroup \em et al.\egroup
  }{2019a}]{xu2019adversarial}
Han Xu, Yao Ma, Haochen Liu, Debayan Deb, Hui Liu, Jiliang Tang, and Anil Jain.
\newblock Adversarial attacks and defenses in images, graphs and text: A
  review.
\newblock {\em arXiv preprint arXiv:1909.08072}, 2019.

\bibitem[\protect\citeauthoryear{Xu \bgroup \em et al.\egroup
  }{2019b}]{xu2019topology}
Kaidi Xu, Hongge Chen, Sijia Liu, Pin-Yu Chen, Tsui-Wei Weng, Mingyi Hong, and
  Xue Lin.
\newblock Topology attack and defense for graph neural networks: An
  optimization perspective.
\newblock {\em arXiv preprint arXiv:1906.04214}, 2019.

\bibitem[\protect\citeauthoryear{Zang \bgroup \em et al.\egroup
  }{2020}]{zang2020graph}
Xiao Zang, Yi~Xie, Jie Chen, and Bo~Yuan.
\newblock Graph universal adversarial attacks: A few bad actors ruin graph
  learning models.
\newblock {\em arXiv preprint arXiv:2002.04784}, 2020.

\bibitem[\protect\citeauthoryear{Z{\"u}gner and
  G{\"u}nnemann}{2019}]{zugner2019adversarial}
Daniel Z{\"u}gner and Stephan G{\"u}nnemann.
\newblock Adversarial attacks on graph neural networks via meta learning.
\newblock {\em arXiv preprint arXiv:1902.08412}, 2019.

\bibitem[\protect\citeauthoryear{Z{\"u}gner \bgroup \em et al.\egroup
  }{2018}]{zugner2018adversarial}
Daniel Z{\"u}gner, Amir Akbarnejad, and Stephan G{\"u}nnemann.
\newblock Adversarial attacks on neural networks for graph data.
\newblock In {\em Proceedings of the 24th ACM SIGKDD International Conference
  on Knowledge Discovery \& Data Mining}, pages 2847--2856. ACM, 2018.

\end{thebibliography}

\end{document}